\documentclass[10pt,twocolumn,letterpaper]{article}

\usepackage{iccv}
\usepackage{times}
\usepackage{epsfig}
\usepackage{graphicx}
\usepackage{amsmath}
\usepackage{amssymb}
\usepackage{multirow}
\usepackage{ulem}
\usepackage{ctable}
\usepackage{adjustbox}
\usepackage{subcaption}

\usepackage[breaklinks=true,bookmarks=false]{hyperref}

\iccvfinalcopy % *** Uncomment this line for the final submission

\def\iccvPaperID{8} % *** Enter the ICCV Paper ID here

\ificcvfinal\fi

\begin{document}

%%%%%%%%% TITLE
\title{Great Ape Detection in Challenging Jungle Camera Trap Footage\\ via Attention-Based Spatial and Temporal Feature Blending\vspace{-8pt}}

\author{Xinyu Yang\\
Dept of Computer Science\\
University of Bristol, UK\\
{\tt\small xinyu.yang@bristol.ac.uk}
\and
Majid Mirmehdi\\
Dept of Computer Science\\
University of Bristol, UK\\
{\tt\small m.mirmehdi@bristol.ac.uk}
\and
Tilo Burghardt\\
Dept of Computer Science\\
University of Bristol, UK\\
{\tt\small tilo@cs.bris.ac.uk}
}

\maketitle
\thispagestyle{empty}

%%%%%%%%% ABSTRACT
\begin{abstract}
%\mmn{We are under the page limit and there are a lot of negative vspaces in the paper so they may be unnecessary. I hope changes to the format won't cause problems} TILO: UNDERSTOOD
\vspace{-8pt} We propose the  first multi-frame video object detection framework  trained to detect great apes. It is applicable to challenging camera trap footage  in complex jungle environments and extends a traditional feature pyramid architecture by adding  self-attention driven feature blending  in both the spatial as well as the temporal domain. We demonstrate that this extension can detect distinctive species appearance and motion signatures despite significant partial  occlusion. We evaluate the framework using $500$~camera trap videos of great apes from the Pan African Programme containing $180K$~frames, which we manually annotated with accurate per-frame animal bounding boxes. These clips contain significant partial occlusions, challenging lighting, dynamic backgrounds, and natural camouflage effects. We show that our approach performs highly robustly and significantly outperforms  frame-based detectors. We also perform detailed ablation studies and a validation on the full ILSVRC 2015 VID data corpus to demonstrate wider  applicability at adequate performance levels. We conclude that the framework is ready to assist human camera trap inspection efforts. We publish key parts of the code as well as network weights and ground truth~annotations with this paper.\vspace{-24pt}
\end{abstract}

%%%%%%%%% BODY TEXT
%-------------------------------------------------------------------------
\section{Introduction } \label{intro}
\begin{figure}[t]\vspace{-10pt}
\includegraphics[width=237pt,height=307pt]{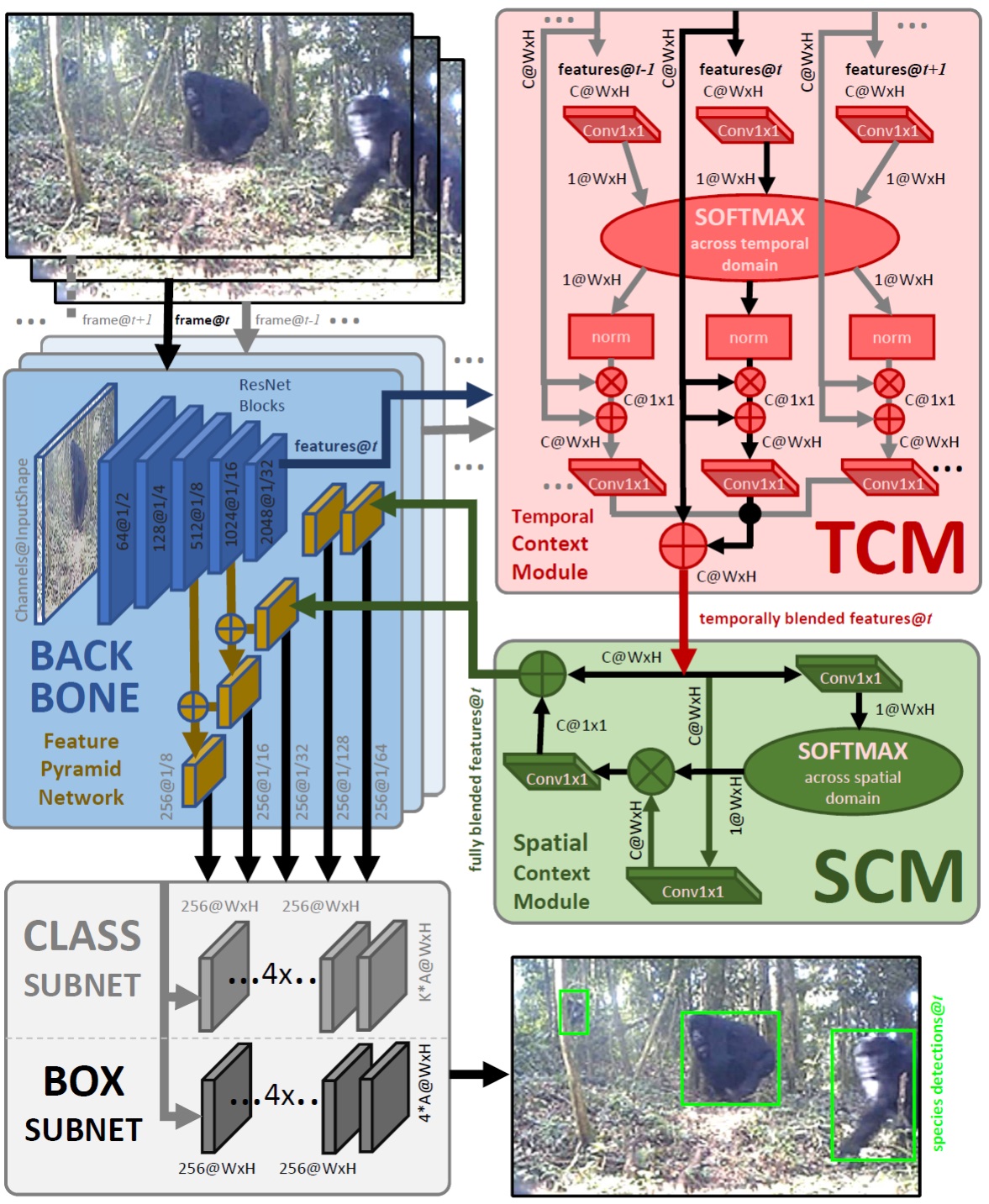}\vspace{-4pt}
  \caption{
  \small{
  \textbf{Architectural Overview.} Our framework processes video snippets (top left) and robustly detects per-frame great ape locations~(bottom right). The architecture extends a traditional detection backbone~(blue) by two self-attention components,~which select and `blend' most relevant temporal~(TCM, red) and spatial (SCM, green) information via SoftMax folding. Blended feature maps are fed back into the feature pyramid network~(yellow) before fuelling the detection head, which contains bounding box regression~(black) and classification~streams~(grey).
    }
  }
  \vspace{-16pt}
\label{fig:arch}
\end{figure}

The problem of visually identifying the presence and locations of animal species filmed in natural habitats~\cite{kuhlburghardt2013} is of central importance for automating the interpretation of large-scale camera trap imagery. This is particularly challenging in scenarios where lighting is difficult, backgrounds are non-static, and major occlusions, image noise,  as well as animal camouflage effects occur: filming great apes via camera traps in jungle environments constitutes one such setting. There, animals appear uniformly dark and blend into the forest during eating, playing or moving in groups sometimes behind trees or thicket. An animal's visual presence and location is thus often only determinable by linking selective spatial and temporal information about species-typical appearance and locomotion across video snippets. 

In this paper, we address this  specific video object detection challenge by proposing a blended detection framework integrated into a feature pyramid network (FPN) architecture \cite{He2016DeepRecognition,Lin2017FeatureDetection} as illustrated in Figure~\ref{fig:arch}.  We introduce  two extra components to a traditional detection backbone:~a~Spatial Context Module (SCM) and a Temporal Context Module (TCM). Each of these modules is driven by a self-attention mechanism  tasked to learn how to emphasise  most relevant elements of a feature given its context. In particular, these attention components are effective in learning how to `blend' spatially and temporally distributed visual cues in order to reconstruct object locations under dispersed partial information; be that due to occlusion or lighting. 

\begin{figure}[b]\vspace{-10pt}
\includegraphics[width=237pt,height=188pt]{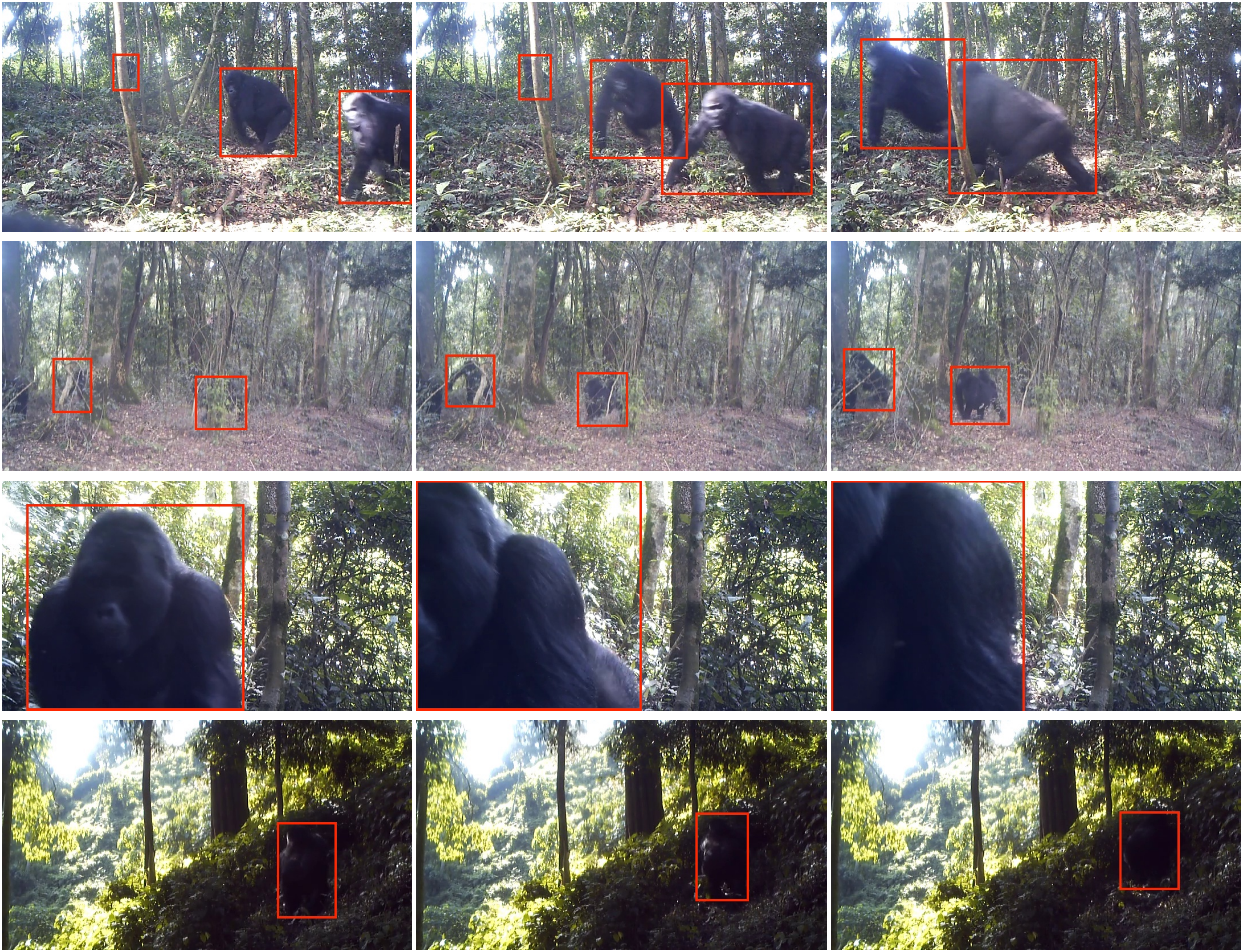}\vspace{-33pt}
\begin{center}
\end{center} %\vspace{-13pt}
  \caption{
  \small{
  \textbf{Pan Africa Great Ape Camera Trap Dataset.}  Sample frames from four representative videos of the dataset with ground truth animal annotations in red. \textbf{\textit{(1$^{st}$ row)}}~clip showing two easily detectable animals walking past a camera trap with a third animal moving behind thicket; \textbf{\textit{(2$^{nd}$ row)}}~clip where two great apes hide and emerge from behind bushes; \textbf{\textit{(3$^{rd}$ row)~}}clip  with frames of extreme close-ups;\textbf{\textit{ (4$^{th}$ row)~}}clip with challenging lighting.
  }
  } %\vspace{-8pt}
\label{fig:dataset}
\end{figure}

Whilst the self-attention concept has been used recently in various application contexts \cite{Wang2018Non-localNetworks,Cao2019GCNet:Beyond}, we tailor it here to spatio-temporal video object detection and propose a flexible component setup that can be utilized as an add-on to different backbone networks.  We show that the approach is beneficial in scenarios where distinctive species appearance and motion signatures  are only partly accessible and intermittently dispersed across the spatial and temporal domain. Figure~\ref{fig:dataset} exemplifies such scenarios on the Pan Africa camera trap data used in this work. This dataset contains~$500\times$ $15s$ video clips with the resolution of $720\times 404$ and was collected by the Pan African Programme `The Cultured Chimpanzee'~(see Acknowledgements). It was subsequently labelled for this paper with accurate bounding box ground truth for all animals in each frame. 

\textbf{Contributions.} \textbf{\textit{(1)}} FPN-integrated blended detection framework driven by spatial and temporal self-attention components;
\textbf{\textit{(2)}} Cross-dataset ablation study of the framework quantifying details for various backbones and setups; \textit{\textbf{(3)}} Comprehensive animal bounding box annotations in 180K frames of the Pan Africa dataset.

%-------------------------------------------------------------------------
\section{Related Work}
Object detection in video is usually implemented via video feature extraction~\cite{Simonyan2014Two-StreamVideos,Carreira2017QuoDataset,Karpathy2014Large-scaleNetworks}, frame feature aggregation~\cite{Chen2018OptimizingLattice,Zhu2017DeepRecognition,Zhu2017Flow-GuidedDetection}, or detection post-processing~\cite{han2016Seq-NMSDetection,Kang2018T-CNN:Videos,Kang2016ObjectNetworks}. The task is distinct from object tracking \cite{Zhu2018Distractor-awareTracking,Li2018SiamRPN++:Networks,Bertinetto2016Fully-convolutionalTracking} since the object instances to follow are not given -- they may appear or disappear at any point and need to be localised and determined as absent or present on a per-frame basis. 

Similar to other video tasks, video object detection relies on extracting relevant information from the spatio-temporal corpus formed by frame sequences.
% \mmn{I am not sure about the end of that sentence. It could have been borrowed from object tracking, person re-id, and many other areas, why action recognition in particular?}. 
For instance, C3D~\cite{Tran2015LearningNetworks} and I3D~\cite{Carreira2017QuoDataset} explore 3D convolutional neural networks~(CNNs) to generate mixed motion-RGB features from frame sequences. In contrast, dual-stream ConvNets such as Siamese ConvNet~\cite{Simonyan2014Two-StreamVideos} and TSN~\cite{Wang2016TemporalRecognition} apply separate networks for RGB and motion processing where optical flow is often used as a pre-computed input. Frame feature aggregation, as used in DFF~\cite{Zhu2017DeepRecognition}, FGFA~\cite{Zhu2017Flow-GuidedDetection}, and ST-Lattice~\cite{Chen2018OptimizingLattice}, deals with motion implicitly by training networks that fuse frame-level spatial features  to explore the temporal structure of videos. Detection post-processing, such as used in T-CNN~\cite{Kang2018T-CNN:Videos} and Seq-NMS~\cite{han2016Seq-NMSDetection}, ignores explicit temporal feature construction altogether. Operating closer to a traditional tracking paradigm, these methods instead link detection results from individual frames into tracklets by optimising overall trajectory scores. 

Before focusing on the most relevant details of some of the above methods, we will first review the foundations on which many of the approaches rest, that is: how to  extract detection-relevant features from single~frames.

%-------------------------------------------------------------------------
\subsection{Single Image Object Detection}
Object detection fundamentally requires two conceptual tasks to be solved: localisation and classification of content of interest. Region proposal based methods such as F-RCNN~\cite{Ren2017FasterNetworks.} or Cascade-RCNN~\cite{Cai2018CascadeDetection} pre-process images first to generate class-agnostic regions of interest (ROIs) before classifying these and regressing associated bounding boxes. F-RCNN, in particular, uses a region proposal network~(RPN) built on top of a backbone shared with the classification component. More recently, to improve the quality of detection, Cascade-RCNN appends two more stages based on F-RCNN and replaces ROI pooling with ROI alignment making the framework more robust. 

In contrast to all region proposal based methods, single-shot detectors infer class probability and bounding box offsets within a single feed forward network. This approach is usually simpler in design and faster at runtime as impressively shown, for instance, by YOLO~\cite{Redmon2016YouDetection,Redmon2017YOLO9000:Stronger} and SSD~\cite{Liu2016SSD:Detector}. Darknet-based YOLO~\cite{Redmon2016YouDetection} in particular  regresses anchor boxes and box scores directly as well as class probabilities. Due to its versatility and fast performance, YOLO has formed the detection backbone for successfully constructed primate face detectors~\cite{Brust2017TowardsWild} for single images in the past. Similarly to the YOLO approach, SSD predicts a fixed number of bounding boxes. However, by applying detection at different scales, SSD has been shown to adjust better to different object sizes and aspect ratios. More recently, by addressing the class imbalance problem of all of the previous single-shot methods, RetinaNet~\cite{Lin2017FocalDetection} replaces the cross-entropy loss with the focal loss for classification, which focusses the training  on hard examples.
%-------------------------------------------------------------------------
\subsection{Video Object Detection}
Whilst single image object detectors are directly applicable to video in a frame-by-frame manner, they ignore -- by definition -- temporal cues. Yet, these are often vital for detection under challenging conditions as for the case at hand. We will next briefly review the key ideas behind most relevant recent works~such as FGFA~\cite{Zhu2017Flow-GuidedDetection}, D\&T \cite{Feichtenhofer2017DetectDetect}, T-CNN~\cite{Kang2018T-CNN:Videos}, and Seq-NMS~\cite{han2016Seq-NMSDetection}. The fundamental task for all these methods is to integrate information from the temporal domain directly into detection decisions.

Linking single frame detections across the temporal dimension as done by T-CNN~\cite{Kang2018T-CNN:Videos} constitutes possibly the simplest form of temporal domain exploration. T-CNN essentially runs region-based detectors per frame and enforces motion-based propagation to adjacent frames. This classical tracking paradigm thereby extends detections into tubelets, which after re-scoring and suppression of overlaps yield the final  set of detected objects. 
In contrast, D\&T \cite{Feichtenhofer2017DetectDetect} interlinks single image detection and tracking in a unified approach using ROI pooling on both detection-based feature maps and tracking-based correlation feature maps, where a specific correlation layer is introduced to produce the latter.

Seq-NMS~\cite{han2016Seq-NMSDetection} follows a similar paradigm constructing sequences along temporally close, high confidence bounding boxes in consecutive frames. 
%% Not sure this is a common way of addressing such things and it stops the reader in their tracks before they cotton on
%There, the 
Their governing metric for sequence association maximises overall confidence and IOU scores, and sequence-based non-maximum suppression is utilised to fuse or filter out overlapping tracklets. Whilst temporal consistency can be extrapolated this way, the spatial distortion effects across the temporal domain are not accounted for. 

To address this, FGFA~\cite{Zhu2017Flow-GuidedDetection} performs optical flow guided spatial warping before aggregating features. The resulting features are subsequently fused temporally by weighted element-wise addition where weights are determined by the optical flow field. These descriptors contain rich spatio-temporal information that have been shown to address problems such as blurred object detection and partially occluded object detection. 

Taking closest inspiration from the temporal cue aggregation mechanism used in FGFA, we propose an attention-based spatial \textit{and} temporal feature blending framework that can be used as an add-on to existing (and future) feature pyramid networks as they are in standard use for object detection today~\cite{Lin2017FocalDetection}. Attention-based spatial blending was successfully used before in  GCNet~\cite{Cao2019GCNet:Beyond}, but attention-based spatio-temporal blending is -- to the best of our knowledge -- novel as a core concept for video object detection. 

We will now describe  how  such information blending across  space and time can be implemented, and how it is beneficial to addressing the problem of great ape detection. 
\begin{figure}[b]
\centering \vspace{-13pt}
\includegraphics[scale=0.37]{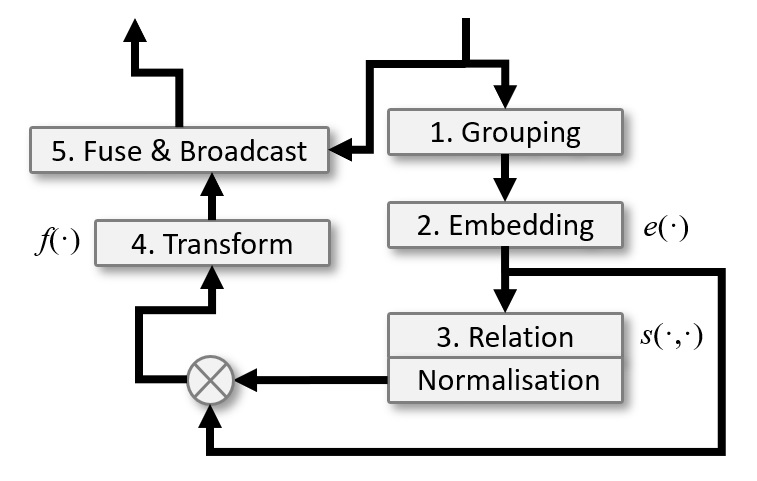}\vspace{-5pt}
  \caption{
  \small{\textbf{Principled Workflow of the Blending Framework.} \textit{\textbf{(1)}} Input features  are grouped by the target dimension; \textit{\textbf{(2)}}~Each location of the feature map is embedded by its position; \textbf{\textit{(3)}}~A  correlation map is computed and reduced. This map is then applied back to the embedded feature via  matrix multiplication (marked as $\bigotimes)$ implementing the self-attention principle.  \textbf{\textit{(4)}}~The resulting feature is finally transformed, before \textbf{\textit{(4)}}~ being fused with the residual to form the blended output. Key functions are shown at the point of application as annotations.}
  }\vspace{-18pt}
\label{fig:concept}
\end{figure}
\vspace{-12pt}

%------------------------------------------------------------------------
\section{Detection via Spatial \& Temporal Blending}
\textbf{Task.} Our task is to generate frame-level bounding box detections~$\mathbf{D}=\{\mathbf{D}_1, \dots, \mathbf{D}_t, \dots\}$ of great apes across a video $\mathbf{X}=\{\mathbf{X}_1, \dots,\mathbf{X}_t,\dots\}$, where $\mathbf{D}_t$ are the detection results for frame $\mathbf{X}_t$ at time step~$t$.  To predict the bounding box detections~$\mathbf{D}_t$, the network can utilise both previous \textit{and} future frames, $i.e.$ a snippet~$\{\mathbf{X}_{t-\tau}, \dots, \mathbf{X}_t, \dots, \mathbf{X}_{t+\tau}\}$ of length~$T$. \vspace{5pt}

\textbf{Overall Concept.} As shown in Figure~\ref{fig:arch}, our integrated architecture extends a standard feature pyramid network by two extra components: a Spatial Context Module~(SCM) and a Temporal Context Module (TCM). Each of these modules is driven by a self-attention mechanism  that learns how to emphasise the most relevant elements of a feature given its context. Both components follow a principled workflow similar to the one described by Cao et al~\cite{Cao2019GCNet:Beyond} and visualised in Figure~\ref{fig:concept}. Essentially, after grouping inputs along the dimensions of attentional interest (spatial or temporal), features are embedded into a lower dimensional space and a self-attention map of the feature is created. This map is then applied back onto the features in order to `blend' it and emphasise  elements important to the detection process whilst suppressing other content. Critically, these components are trainable as part of the network and can be rolled  out across space and time so that dispersed species-distinctive information can be selected from within the spatio-temporal volume. 

For a given spatial or temporal module and position~$i$ in the input feature~$x_i$, the context-enhanced and `blended' output feature~$z_i$ can in its simplest form be expressed as: \vspace{-6pt}
\begin{equation}
\label{equ1}
    z_i=x_i+f\bigg(\frac{1}{M}\sum^{M}_{j=1} s\Big(e_i(x_i),e_j(x_j)\Big) e_i(x_i)\bigg) ~, \vspace{-4pt}
\end{equation}
where $x_i$ is {the} $i$th descriptor of the residual feature map, $e(\cdot)$~is the embedding function, $f(\cdot)$ is the transform function, $s(\cdot)$~is the correlation function, $M$ is the number of positions in the feature, $e.g.$ $M=HWT$ for {a} video {sequence} {($H$ denotes height, $W$ denotes width, $T$ denotes snippet length)}, and  $j$~enumerates all the positions across the context. The embedding function $e(\cdot)$ and transfer function $f(\cdot)$ are  implemented via $1\times 1$ convolution kernels without bias using learnable weights $w$.\vspace{5pt}

% \mmn{we have so far always said spatial and temporal, or even spatiotemporal, so ideally the position of the two modules next should be swapped.}
\textbf{Spatial Module.} As depicted in Figure~\ref{fig:arch} (green), we use a simplified non-local attention component with a SoftMax function applied across the embedded feature vector as  the SCM. The module output $z_i^{SCM}$ is:\vspace{-5pt}
\begin{small}\begin{equation}
    z_{i}^{SCM}=x_{i} \oplus w^3 \sum_{j=1}^{M}\left(\begin{small}\frac{\exp{(w^1x_j)}}{\sum^{M}_{a=1}\exp{(w^1x_a)}}\end{small}\otimes(w^2x_j) \right) ~, \vspace{-3pt}
\end{equation}\end{small}
where  $j$ enumerates all  context locations, $x_i$ represents the incoming features, $\oplus$ denotes element-wise broadcast addition, $\otimes$ denotes multiplication of tensor elements, $w^{1}$~and  $w^{2}$~describe the different learnable parameters of linear embedding functions $e(\cdot)$, and  $w^{3}$~represent the parameters of a linear transform function $f(\cdot)$.

%\mmn{The use of $w$ in the above and below is not very clean or consistent?} TILO: DONE

\begin{figure}[b]
\begin{center}
\includegraphics[width=237pt,height=180pt]{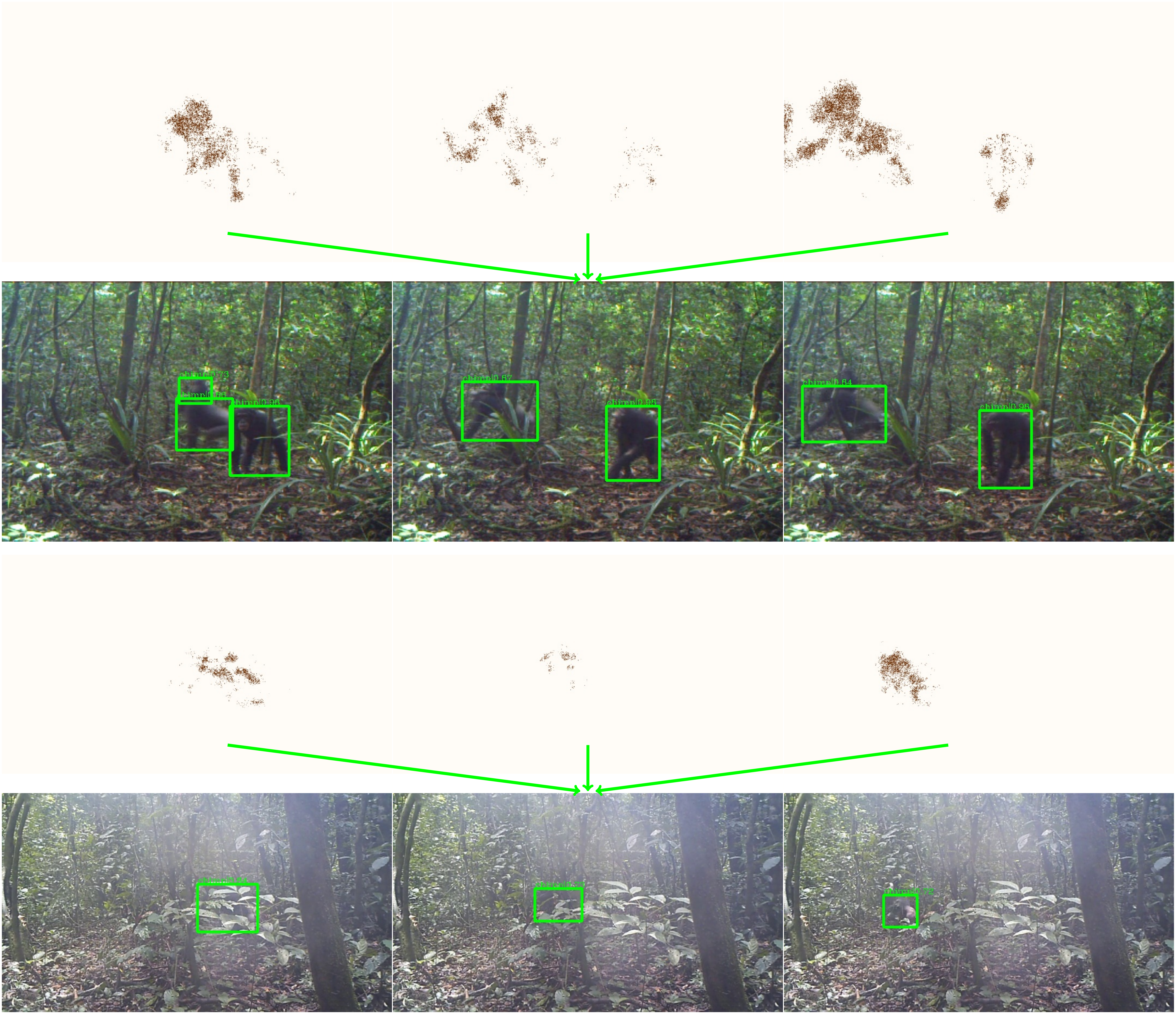}
\end{center}\vspace{-13pt}  \caption{\small{\textbf{Visualisation of Temporal Attention Maps for Two Example Scenes.} First and third rows depict attention maps when projected back into the image domain.  Second and fourth row show the associated frames with green bounding boxes showing predictions by the network with $T=3$. Green lines indicte that  shown attention maps contribute to detections in the central frame.}}\vspace{-10pt}% explain figure 5 here
\label{fig:attention maps}
\end{figure}

\vspace{5pt}
\textbf{Temporal Module.} Figure~\ref{fig:arch} (red) visualises the temporal module in detail, which follows the principles layed out in Equation~(\ref{equ1}) in general terms. In particular, the TCM module is constructed to apply self-attention across a short-term temporal context where, for each input frame $X_t$, a feature $x_t$ from the previous layer is first embedded by a linear function~$e(\cdot)$ with weights $w^4_{t}$. Subsequently, its  temporal correlation function $s(\cdot)$ is modelled by a global SoftMax  $\mathcal{C}(\cdot)$ across a temporal context of $T=2\tau+1$ selected nearby frames. For each feature position $i \in M$ the global SoftMax is defined as:
\vspace{-12pt}
\begin{equation}
    \mathcal{C}(x_{t,i} ; w^4)=\frac{\exp{(w^{4}_{t}x_{t,i})}}{\sum_{m \in T}\exp{(w^4_{m}x_{m,i})}} ~, \vspace{-2pt}
\end{equation}{}
where frame $m$ enumerates all $T$ frame positions, and $w_{t}$ is the linear embedding parameter for time step $t$. We then use the mean value of all positional temporal attention to normalise the term yielding a temporal attention  map for time step $t$ as:\vspace{-12pt}
\begin{equation}
    \hat{x}_{t,i}=\frac{1}{HW}\mathcal{C}(x_{t,i} ; w^4)\sum_{j=1}^{HW}\mathcal{C}(x_{t,j} ; w^4) ~.\vspace{-3pt}
\end{equation}
In order to visually illustrate this concept, two examples of temporal attention maps projected back into the image domain are depicted in Figure~\ref{fig:attention maps}. The maps highlight distinctive, dispersed features for context-aware inference revealing target object locations despite heavy partial occlusion.

These maps are subsequently applied back onto the original feature by matrix multiplication~$\otimes$, element-wise broadcast addition~$\oplus$, and two linear transformations. One transform learns adaptive weights (\textit{i.e.} $w^5_{m}$) to `blend' between original and attention-adapted features, the other learns weights  (\textit{i.e.} \textit{$w^6_{m}$}) to `blend' across the temporal domain. Finally, an additive connection to the original inputs~$x_{t,i}$ is made to ensure learning stability yielding: \vspace{-8pt}
\begin{equation}
    z_{t,i}^{TCM}=x_{t,i}+\sum_{m \in T}w^6_{m}(x_{m,i} \oplus w^5_{m}\sum_{j=1}^{HW}\hat{x}_{m,j}\otimes x_{m,j})  ~.\vspace{-3pt}
\end{equation} \vspace{5pt}

%%% Hi Xinyu, please don't label figures with figureX, as they often will not end up as such. Name them more informatively.
%%% sure, no problem.

%\mmn{Should we not mention the form of the embedding function $e(.)$, and less importantly, have a superscript on $z$ to indicate spatial or temporal in eqs. (4) and (5)}? TILO: DONE

%-------------------------------------------------------------------------
\vspace{-9pt}
\subsection{ Overall Network Architecture}
\textbf{Backbones.} {We follow the layout of   RetinaNet~\cite{Lin2017FocalDetection}} and Cascade~\cite{Cai2018CascadeDetection} as the {two major} state-of-the-art architectures in our study.  Figure~\ref{fig:arch}~(blue) visualises a ResNet50 architecture as an example backbone \cite{He2016DeepRecognition} where we extract FPN levels from network layers $P_3$ to~$P_7$. Across different scales of the FPN, anchors cover area sizes of $32^2$ to $512^2$ distributed over pyramid levels from $P_3$ to $P_7$, respectively. 
We embed our TCM and SCM after the  last layer of the backbone for a maximal receptive field during blending. \vspace{4pt} 

\textbf{Detection Head.} The RetinaNet based detection head contains two streams, \textit{i.e.} a localisation stream and classification stream. The classification subnet (coloured grey in Figure~\ref{fig:arch}) predicts a likelihood vector of class membership for each anchor box. This network is fully convolutional and operates on top of the $5$ FPN layer outputs $P_3$ to $P_7$.  It is composed of $4$ cascaded $3 \times 3$ convolutional layers, each with $256$ channels, and followed by a ReLU activation. The final layer uses $3 \times 3$ convolutions with a Sigmoid function as activation for producing the final classification output. In the localisation subnet (coloured black in Figure~\ref{fig:arch})  the final layer yields  $4\times$ channel outputs describing bounding box locations. We assign initial anchor areas ranging from $32^2$ to $512^2$ elements corresponding to the pyramid levels $P_3$ to $P_7$, respectively. Subsequent rescaling uses the ratios of [$2^0$, $2^{1/3}$, $2^{2/3}$] and transforms anchors with three aspect ratios [1:2, 1:1, 2:1]. In total there are 9 anchors per level. Focal loss from \cite{Lin2017FocalDetection} is adopted to tackle the foreground and background imbalance problem.

We also implemented a stronger, cascaded detection head~\cite{Cai2018CascadeDetection} for maximum detection quality. The cascaded detection head contains four stages, one for an RPN, which regresses class-agnostic regions of interest, and three for detection with different IoU ground truth assignments parameterized as~$[0.5,0.6,0.7]$. Note that the three cascaded detection heads are fully convolutional and ROI alignment is used for ROI feature pooling. 

%------------------------------------------------------------------
\subsection{Training Details}
We train the network end-to-end in two separate stages~(see Figure~\ref{fig:training}). First, we pre-train our model on the ImageNet VID dataset for 14 epochs, and then the entire model is trained with synchronised batch normalisation and {training data augmentation (sequence-based random brightness, horizontal flip, central crop)} for $14$ epochs on the $400$ video training portion of the Pan Africa dataset. The remaining $100$ clips are used for testing. 

During the various experiments, we use short-term video snippets and sparse sampling of between $T=3$ to $T=8$ frames {whose shorter size is }resized {to at least $512$ and up to $800$} pixels horizontally keeping the aspect ratio constant by padding if necessary. During training, each mini-batch contains $8$ snippets, that is $1$ per deployed Nvidia Tesla P100 GPU.  The learning rate is warmed up for the first $500$ iterations of each training phase from $0.002$ to $0.01$ and decreased $10$ times at epoch $6$ and $11$, respectively. 

We use SGD with the momentum of $0.9$ as the optimizer and utilise ImageNet pre-trained weights to initialise the backbone. FPN layers and all other convolutional layers are initialised by normal distributions.\vspace{-5pt}

%-------------------------------------------------------------------------
\section{Experimental Results}
We evaluate the proposed system both quantitatively and qualitatively, and perform detailed ablation studies. We also perform  a validation on the full ILSVRC 2015 VID data corpus to demonstrate wider  applicability.

%\mmn{Sorry, to speed up, I will from now on just edit without colouring unless it is absolutely essential. Edits are visible in the late as I have turned tracking on.} TILO: OK

\subsection{Quantitative Evaluation}
Using the test portion of the Pan Africa dataset, we first evaluate  single frame detectors as our baselines. For evaluation, we compute the average precision (AP) as the area under precision-recall curve and then report the mean of AP~(mAP) for the classes in the validation set. 

We re-implemented RetinaNet~\cite{Lin2017FocalDetection} with ResNet50 and ResNet101 baselines and, as shown in Table \ref{tab:chimpresult}, these two architectures achieve 80.79\% and 85.25\% mAP, respectively. Whilst adding a SCM component to the ResNet50 RetinaNet setup improves performance slightly to 81.21\%, the current state-of-the-art Cascade-RCNN~\cite{Cai2018CascadeDetection} outperforms this setup and older single frame baselines, achieving 88.31\%  on the Pan Africa test data. 
The basic addition of a TCM component, on the other hand, produces significant performance improvements by up to $10.02\%\uparrow$ for ResNet50 RetinaNet, and still $2.86\%\uparrow$ for Cascade-RCNN. 

\begin{figure}[t]
\includegraphics[width=237pt,height=111pt]{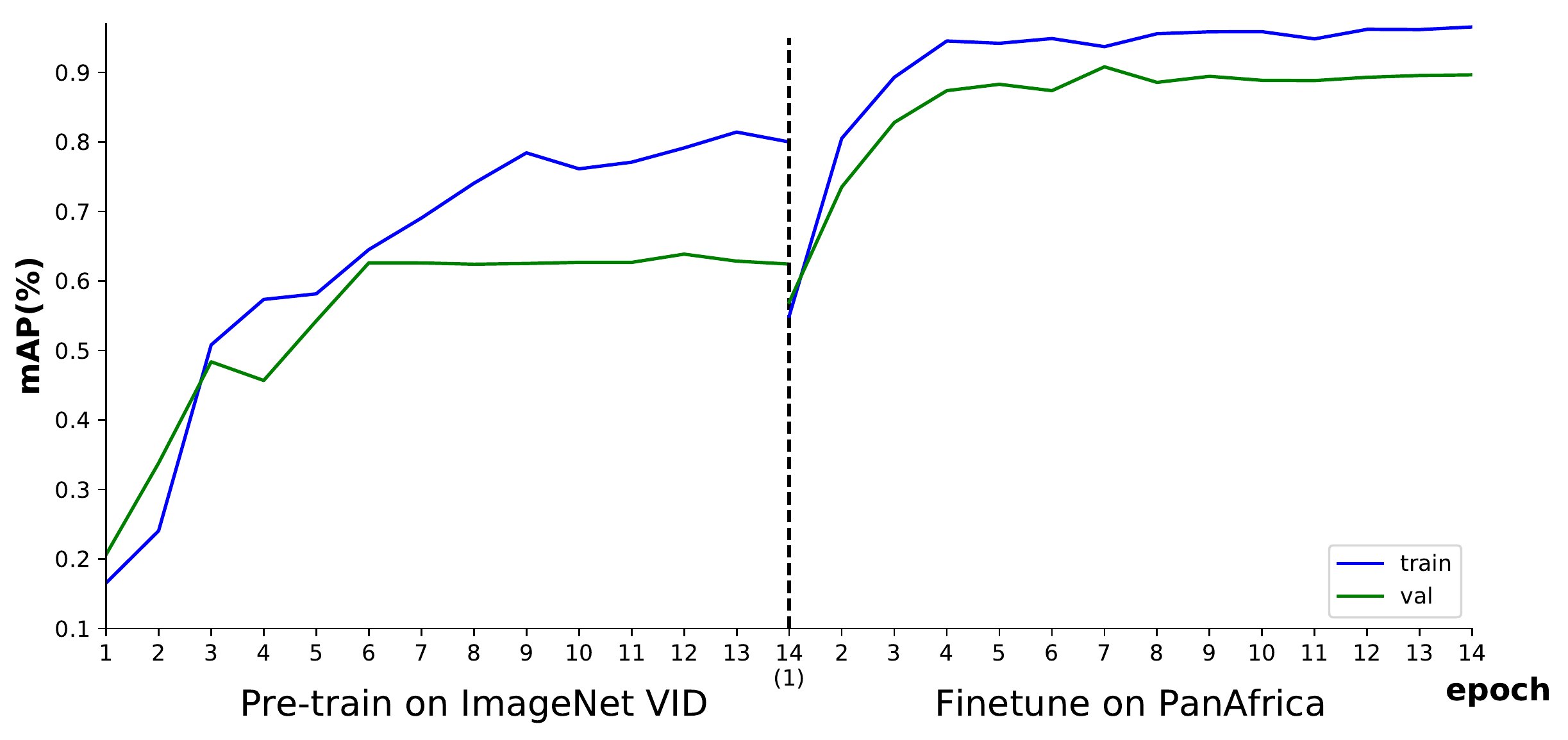}\vspace{-5pt}
  \caption{
  \small{
  \textbf{Example Training and Validation Performance.}  {mAP plots} for both training (blue) and validation~(green)  of the {ResNet50} RetinaNet SCM+TCM setup along the $28$ epochs of optimisation split into two distinct phases, i.e. pre-training on ImageNet VID, and fine-tuning on the Pan Africa dataset. }  %% slight change to reduce the caption by a line
  }\vspace{-4pt}
\label{fig:training}
\end{figure}

\begin{table}[b]
\begin{adjustbox}{width=\columnwidth,center}
\begin{tabular}{cc|cc|c}

\multicolumn{2}{c|}{Backbone and FPN Layout} & $T_{train}$   & $T_{test}$     & mAP(\%) \\
\hline
\begin{tabular}[c]{@{}l@{}}Res50\\ RetinaNet\end{tabular}  & \begin{tabular}[c]{@{}l@{}}baseline\cite{Lin2017FocalDetection}\\ +SCM\\ +TCM\\ +SCM+TCM\end{tabular} & \begin{tabular}[c]{@{}l@{}}\textbackslash\\ \textbackslash\\ 7\\ 7\end{tabular} & \begin{tabular}[c]{@{}l@{}}\textbackslash\\ \textbackslash\\ 21\\ 21\end{tabular} & \begin{tabular}[c]{@{}l@{}}80.79\\81.21\\90.02\\ \textbf{90.81}\end{tabular} \\
\hline
\begin{tabular}[c]{@{}l@{}}Res101\\ RetinaNet\end{tabular} & \begin{tabular}[c]{@{}l@{}}baseline\cite{Lin2017FocalDetection}\\ +SCM+TCM\end{tabular}               & \begin{tabular}[c]{@{}l@{}}\textbackslash\\ 5\end{tabular}                      & \begin{tabular}[c]{@{}l@{}}\textbackslash\\ 21\end{tabular}                       & \begin{tabular}[c]{@{}l@{}}85.25\\\textbf{90.21}\end{tabular}              \\
\hline
\begin{tabular}[c]{@{}l@{}}ResX101\\ Cascade\end{tabular} & \begin{tabular}[c]{@{}l@{}}baseline\cite{Cai2018CascadeDetection}\\ +SCM+TCM\end{tabular}               & \begin{tabular}[c]{@{}l@{}}\textbackslash\\ 3\end{tabular}                      & \begin{tabular}[c]{@{}l@{}}\textbackslash\\ 21\end{tabular}                       & \begin{tabular}[c]{@{}l@{}}88.31\\ \textbf{91.17}\end{tabular}             
\end{tabular}
\end{adjustbox}
\caption{\textbf{Pan Africa Performance Results}. Boosts in mean average precision (mAP) on the Pan Africa test dataset when applying the proposed TCM \& SCM components with various state-of-the-art FPN architectures.}\vspace{-8pt}
\label{tab:chimpresult}
\end{table}

The training process for a full SCM+TCM setup, as quantified earlier in Figure~\ref{fig:training} for instance, furthermore reveals that the generalisation gap~of such a model (mAP distance between blue and green curves) narrows significantly for the PanAfrica dataset compared to the ImageNet VID pre-training in late stages. Narrow generalisation gaps indicate that a model is particularly capable of carrying over learned inference strategies  to unseen instances of the particular dataset. For the case at hand, this and the improved overall mAP indicate that fine-tuning is indeed successful.

We also validated a basic ResNet50 RetinaNet version that uses SCM+TCM on the full ILSVRC 2015 VID corpus to show that our rather simple and flexible SCM+TCM network extension can achieve {strong} results also in this general detection setting at $63.85\%$~mAP.  This %setup achieves on this data corpus which would 
ranks within the top three of the original mAP based competition~\cite{ImageNet}.

\begin{figure}[t]
\centering
\includegraphics[width=240pt,height=260pt]{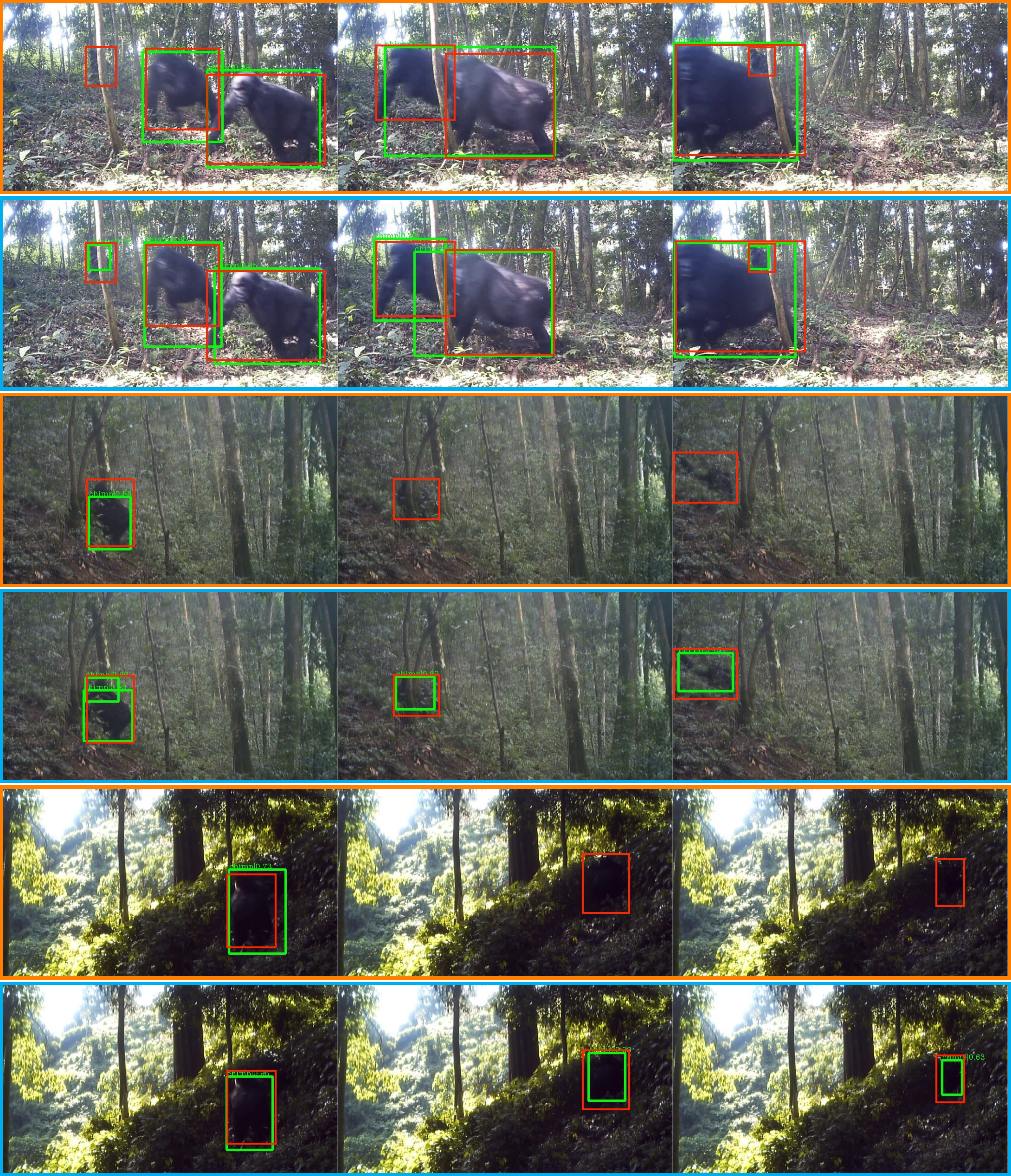}
\vspace{-20pt}
\caption{\small{\textbf{ Examples of  Positive Attention Component Impact.} Depicted are ground truth labels (red) and detections~(green) for a ResNet50 FPN with (rows 2, 4, and 6) and without (rows 1, 3, and 5) SCM+TCM components. Note the clear detection improvements for partially occluded great apes (top two examples) and under challenging lighting (bottom example).}}\vspace{-9pt}
\label{fig:results}
\end{figure}

\subsection{Qualitative Observations on Pan Africa Data} 
\begin{figure}[b]\vspace{-10pt}
\centering
\includegraphics[width=240pt,height=202pt]{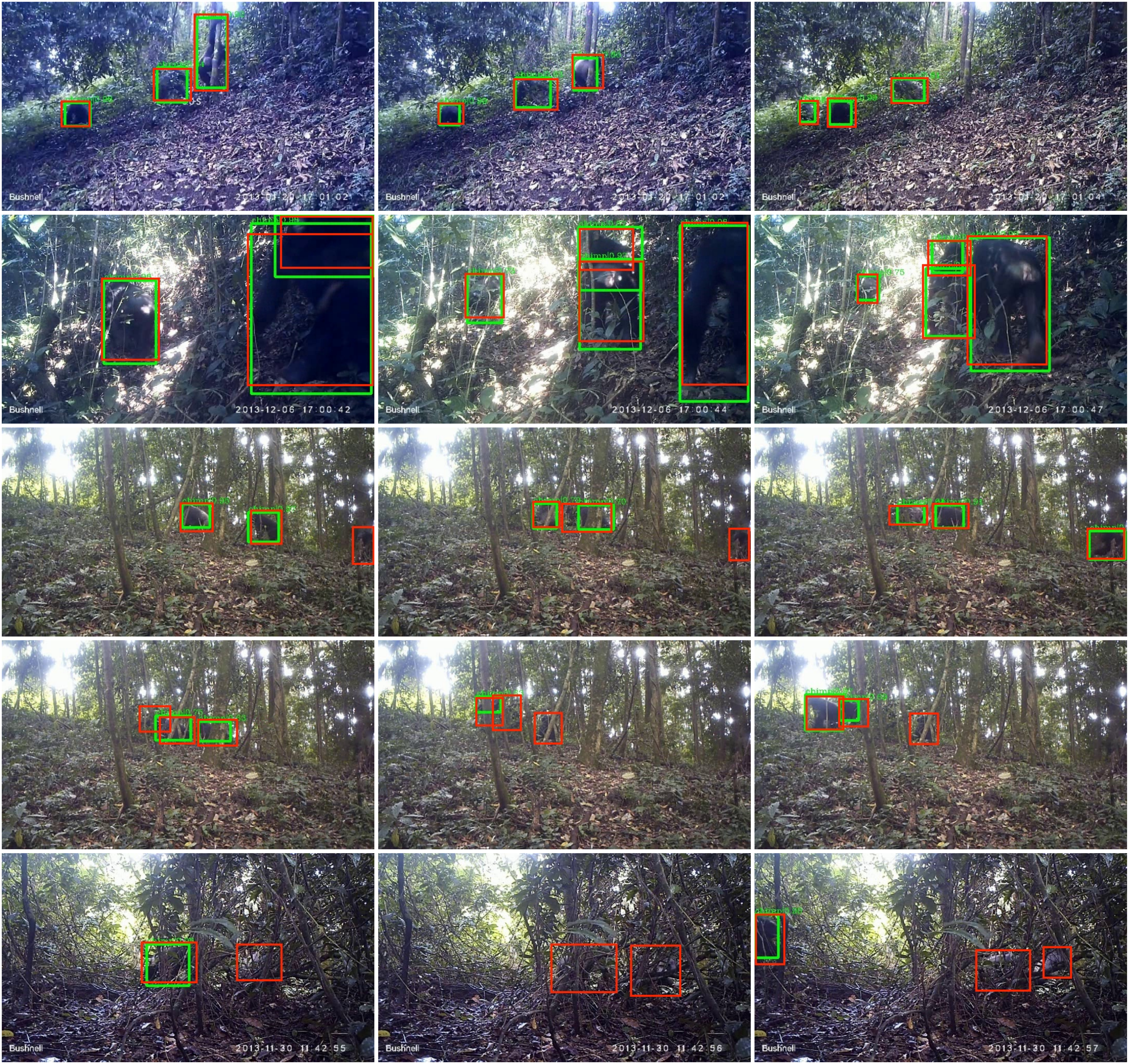}
\vspace{-29pt}
\begin{center}
\end{center}\vspace{-13pt}  \caption{\small{\textbf{Success and Failure Cases.} Depicted are ground truth labels (red) and detections~(green) in frames of various example clips using the best performing model ResX101\ Cascade with  SCM+TCM components.  }}\vspace{-10pt}
\label{fig:results2}
\end{figure}

The Pan Africa dataset contains many scenes where illumination, occlusions, noise or animal camouflage effects make it challenging to recognise animals (as seen before in Figure \ref{fig:dataset}). We found that the SCM+TCM setup consistently improves detection robustness compared to baselines in such cases. These improvements make a significant contribution to the overall quantitative results reported before. Figure~\ref{fig:results} provides  examples of successful cases where per-frame accurate animal detection is achieved by the SCM+TCM components in the presence of partial occlusion or challenging lighting. However, as depicted in the bottom examples of Figure~\ref{fig:results2}, a number of particular animal appearances remain challenging to detect despite the availability of spatial and temporal context information.

%-------------------------------------------------------------------------
\subsection{Ablation Study}
We conduct an ablation study on both the Pan Africa   and the ILSVRC 2015 VID datasets in order to quantify in detail the impact of the key system parameters.  All of the results are reported based on ResNet50+SCM+TCM setup with a RetinaNet detection head.

\textbf{SCM+TCM\ Insertion Point.} Considering arguments in \cite{Cao2019GCNet:Beyond}, we acknowledge that the embedding position of our module along the original backbone network is important. To determine the best insertion point, we test three possible positions for insertion into ResNet50: after the $3 \times 3$ convolution of the last ResNet block; after the last  $1 \times 1$ convolution before the residual addition; and after the residual addition.
As shown in Table \ref{tab:ablation1}(a) and Table \ref{tab:ablation2}(a), we find that the last option is superior and can gain $6.01\%\uparrow$ mAP on Pan Africa and $6.42\% \uparrow$ mAP on VID improvement when compared with insertion after the last $1 \times 1$ convolution. This indicates that the final residual addition
in the base network \textit{does} provide useful extra information.

\textbf{Temporal Frame Support.} Different choices on supporting frames~$T$ are ablated in Table \ref{tab:ablation1}(b) and \ref{tab:ablation2}(b). Results confirm that wider windows for temporal integration do indeed benefit detection, however, particularly during training, GPU sizes limit possible choices of $T$. During testing, longer exposures are possible. Quantifying the effect of varying test exposures we find that, for $T_{test}=21$ frames compared to $T_{test}=5$, there is a  $1.96\% \uparrow$ and $3.69\% \uparrow$ improvement for fixed   $T_{train}=7$ on the Pan Africa and ILSVRC 2015 VID datasets, respectively.\vspace{10pt}

\textbf{Embedding Strategy in the TCM.} We found that when applying an embedding strategy in the TCM where only the current main frame features (indicated as black arrows of the TCM in Figure~\ref{fig:arch}) are backpropagated and reference features are used, but not backprogagated, then a marginally improved  overall performance can be observed (marked as \textit{Main \& Refs} in contrast to \textit{Positional} in Tables \ref{tab:ablation1} and \ref{tab:ablation2}).
  
\begin{table}[b]
\centering

\begin{subtable}{.5\textwidth}
    \caption{\textbf{Position}}\vspace{-2mm}
    \begin{tabular}{ccccc}
    Embedding  & position  &$T_{train}$ & $T_{test}$ & mAP(\%) \\
    \hline
    Positional & after add & 8          & 8         & 88.21   \\
    Positional & after 1x1 & 8          & 8         & 82.20   \\
    Positional & after 3x3 & 8          & 8         & \textbf{87.75}  \\
    \specialrule{.1em}{.05em}{.05em}
    \end{tabular}
\end{subtable}

\begin{subtable}{.5\textwidth}
    \caption{\textbf{Supporting Frames}}\vspace{-2mm}
    \begin{tabular}{ccccc}
    Embedding    & position  & $T_{train}$ & $T_{test}$  & mAP(\%) \\

    \hline
    Main \& Refs & after add & 7          & 21        & \textbf{90.81}   \\
    Main \& Refs & after add & 7          & 5         & 88.85  \\
    Main \& Refs & after add & 3          & 5         & 87.76  \\
    \specialrule{.1em}{.05em}{.05em}
\end{tabular}
\end{subtable}

\begin{subtable}{.5\textwidth}
    \caption{\textbf{Embedding Strategy}}\vspace{-2mm}
\begin{tabular}{ccccc}
Embedding     & position  & $T_{train}$ & $T_{test}$  & mAP(\%) \\

Positional & after add & 7          & 21        & 88.61   \\
Main \& Refs  & after add & 7          & 21        & \textbf{90.81}   \\
\specialrule{.1em}{.05em}{.05em}
\end{tabular}

\end{subtable}
\caption{\textbf{Pan Africa Dataset Ablation Study.}  Key system design choices tested on a ResNet50+SCM+TCM setup with a RetinaNet detection head.}
\label{tab:ablation1}
\end{table}

\begin{table}[b]
\centering

\begin{subtable}{.5\textwidth}
    \caption{\textbf{Position}}\vspace{-2mm}
    \begin{tabular}{ccccc}
    Embedding  & position  & $T_{train}$ & $T_{test}$  & mAP(\%) \\
    \hline
    Positional & after add & 8          & 8         & 56.25   \\
    Positional & after 1x1 & 8          & 8         & 49.83   \\
    Positional & after 3x3 & 8          & 8         & \textbf{58.20}  \\
    \specialrule{.1em}{.05em}{.05em}
    \end{tabular}
\end{subtable}

\begin{subtable}{.5\textwidth}
    \caption{\textbf{Supporting Frames}}\vspace{-2mm}
    \begin{tabular}{ccccc}
    Embedding    & position  & $T_{train}$ & $T_{test}$  & mAP(\%) \\
    \hline
    Positional   & after add & 8          & 8         & 56.25   \\
    Positional   & after add & 5          & 8         & 54.38   \\
    \hline
    Main \& Refs & after add & 7          & 21        & \textbf{63.85}   \\
    Main \& Refs & after add & 7          & 5         & 60.16  \\
    \specialrule{.1em}{.05em}{.05em}
\end{tabular}
\end{subtable}

\begin{subtable}{.5\textwidth}
    \caption{\textbf{Embedding Strategy}}\vspace{-2mm}
\begin{tabular}{ccccc}
Embedding     & position  & $T_{train}$ & $T_{test}$  & mAP(\%) \\
\hline
Positional & after add & 7          & 21        & 59.25   \\
Main \& Refs  & after add & 7          & 21        & \textbf{63.85}   \\
\specialrule{.1em}{.05em}{.05em}
\end{tabular}

\end{subtable}
\caption{\textbf{ILSVRC 2015 VID Dataset Ablation Study.} Key system design choices tested on a ResNet50+SCM+TCM setup with a RetinaNet detection head.}\vspace{-15pt}
\label{tab:ablation2}
\end{table}

\section{Conclusion and Implications}
In this paper we proposed the first multi-frame video object detection framework trained and evaluated for detecting great apes utilising their full body morphology. We demonstrated that the framework is applicable to challenging camera trap footage taken in complex jungle environments.  %Methodologically, we 
We introduced two self-attention driven feature blending  components operating in both the spatial and the temporal domains to facilitate detection under heavy partial occlusions and challenging lighting variations. 

We showed that this novel and flexible extension performs robustly at $91.17\%$~mAP on a real world Pan Africa camera trap $500$ video dataset, which we labelled accurately with animal ground truth annotations for $180$K  frames. We conducted detailed ablation studies on our method and showed that the setup  significantly outperforms  state-of-the-art frame based detectors. For general evaluation beyond the task at hand, we  also performed  a validation on the  ILSVRC 2015 VID data corpus to demonstrate {significant} performance on non-specialised video object detection. 

We note that currently ecological camera trap studies are widely conducted by manual inspection, although great ape face detection~\cite{Brust2017TowardsWild,loos2013automated} has been used for ecological surveys before \cite{crunchant2017automated} and DrivenData \cite{Competition:Factorization} hosted a recent challenge to classify jungle camera trap clips by species, without detecting animals and their location in frames explicitly. 

The presented system, in contrast, provides explicit animal locations and is independent of visibility constraints regarding the animal's face. It adds a new capability of detection and localisation of animals partly occluded by vegetation at adequate performance levels.
Whilst tests against other current video detection frameworks are outstanding and will form part of our future work, we conclude that the presented system is ready to assist human camera trap inspection efforts. 

%-------------------------------------------------------------------------
\vspace{-5pt}\section*{Acknowledgements}
We would like to thank the entire team of the Pan African Programme: `The Cultured Chimpanzee' and its collaborators for allowing the use of their data for this paper. Please contact the copyright holder Pan African Programme at http://panafrican.eva.mpg.de to obtain the dataset. Particularly, we thank: H Kuehl, C Boesch, M Arandjelovic, and P Dieguez. We would also like to thank: K Zuberbuehler, K Corogenes,
E Normand,  V Vergnes, A Meier, J Lapuente,
D Dowd, S Jones,
V Leinert,
E Wessling, H Eshuis,
K Langergraber, S Angedakin,
S Marrocoli,
K Dierks, T C Hicks, J Hart,
K Lee,
and M Murai.
Thanks also to the team at https://www.chimpandsee.org.
The work that allowed for the collection of the dataset was funded by the Max Planck Society, Max Planck Society Innovation Fund, and Heinz L. Krekeler. In this respect we would also like to thank: Foundation
MinistÃ¨re de la Recherche Scientifique, and MinistÃ¨re des Eaux et ForÃªts in Co´te d'Ivoire; Institut Congolais pour la Conservation de la Nature and MinistÃ¨re de la Recherche Scientifique in DR Congo; Forestry Development Authority in Liberia;  Direction des Eaux, ForÃªts Chasses et de la Conservation des Sols, Senegal; and Uganda National Council for Science and Technology, Uganda Wildlife Authority, National Forestry Authority in Uganda.
{
\small
\bibliographystyle{ieee_fullname}
\bibliography{references_}
}
\end{document}